\documentclass{article}
\usepackage{vmargin}
\setpapersize{USletter}
\setmarginsrb{25mm}{25mm}{25mm}{18mm}{0mm}{0mm}{7mm}{7mm}

\usepackage{amsmath,amssymb,amsthm,latexsym}
\usepackage{algorithmicx}
\usepackage{algorithm}
\usepackage[noend]{algpseudocode}
\usepackage{enumerate}
\usepackage{graphics}
\usepackage{graphicx}
\usepackage{xspace}
\usepackage{multirow,adjustbox,listings}
\usepackage{url}
\usepackage[x11names,svgnames]{xcolor}
\usepackage{natbib}
\usepackage{tikz}
\usepackage{subfigure,wrapfig}

\newcommand{\comment}[1]{{}}

\newtheorem{Lem}{Lemma}

\newtheorem{Thm}[Lem]{Theorem}

\newtheorem{Def}{Definition}

\newtheorem{Ex}{Example}

\newcommand{\id}[1]{\mbox{\it #1\/}}

\newcommand{\kw}[1]{\mbox{\tt #1}}

\newcommand{\cond}[2]{\ensuremath{%
\mathtt{prob}(#1 \mid #2)}}

\def\p@enumiii{\theenumi(\theenumii)}

\newcommand\annotate[1]%
{\textcolor{blue}{$^\maltese$}%
\-\marginpar[\raggedleft\scriptsize \textcolor{red}{#1}]%
{\raggedright\scriptsize \textcolor{red}{#1}}}

\begin{document}
\date{}

\title{Adaptive MCMC-Based Inference in Probabilistic~Logic~Programs}
\author{Arun Nampally,  C.\ R.\ Ramakrishnan\\
Department of Computer Science, 
Stony Brook University, 
Stony Brook, NY 11794 \\
\url{{anampally, cram}@cs.stonybrook.edu}\\
}

\maketitle


\begin{abstract}
Probabilistic Logic Programming (PLP) languages enable programmers to specify systems that combine logical models with statistical knowledge.  The inference problem, to determine the probability of query answers in PLP, is intractable in general, thereby motivating the need for approximate techniques.  In this paper, we present a technique for approximate inference of conditional probabilities for PLP queries.  It is an Adaptive Markov Chain Monte Carlo (MCMC) technique, where the distribution from which samples are drawn is modified as the Markov Chain is explored.  In particular, the distribution is progressively modified to increase the likelihood that a generated sample is consistent with evidence.  In our context, each sample is uniquely characterized by the outcomes of a set of random variables.  Inspired by reinforcement learning, our technique propagates rewards to random variable/outcome pairs used in a sample based on whether the sample was consistent or not.  The cumulative rewards of each outcome is used to derive a new ``adapted distribution'' for each random variable.  For a sequence of samples, the distributions are progressively adapted after each sample.  For a query with ``Markovian evaluation structure'', we show that the adapted distribution of samples converges to the query's conditional probability distribution.  For Markovian queries, we present a modified adaptation process that can be used in adaptive  MCMC as well as adaptive independent sampling.  We empirically evaluate the effectiveness of the adaptive sampling methods for queries with and without Markovian evaluation structure.
\end{abstract}

\section{Introduction}
\label{sec:intro}


Probabilistic Logic Programming (PLP) covers a class of Statistical Relational
Learning frameworks~\cite{getoor2007introduction} aimed at combining logical and
statistical reasoning.  Examples of languages and systems combining logical and
statistical inference include ICL~\cite{poole1997independent},
SLP~\cite{muggleton1996stochastic}, PRISM~\cite{Sato1997prism},
LPAD~\cite{vennekens2003general} and ProbLog~\cite{de2007problog}.  \comment{At
  a high level, all these languages add syntactic constructs to specify
  probability distributions to traditional logic programming language.}
\comment{Of these, ICL, PRISM, LPAD, and ProbLog all have a \emph{distribution
    semantics}, which is a declarative semantics defined in terms of a
  probability distribution over models of the underlying non-probabilistic
  program.  } \comment{In related work?  The declarative semantics allows for
  many traditional logic program transformations (e.g. unfolding) to be meaning
  preserving in the probabilistic setting as well.  This is in sharp contrast to
  other probabilistic programming languages such as Church~\cite{} whose
  semantics is specified only operationally.  } In addition to standard
statistical models, these languages allow reasoning over many models where
logical and statistical knowledge is intricately combined, and cannot be
expressed as standard statistical models.

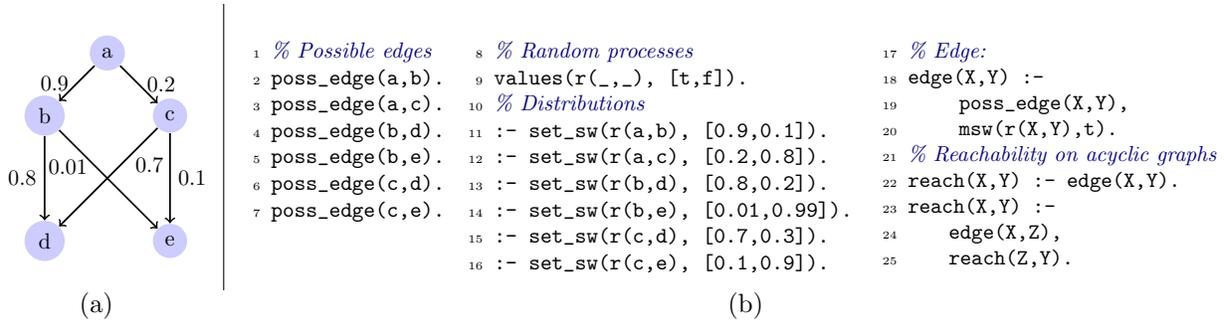
\begin{figure}
  \centering
  \begin{tabular}{cc}
\begin{adjustbox}{width=1in,keepaspectratio}
    \begin{minipage}{1.2in}
      \begin{tikzpicture}
        \tikzstyle{vertex}=[circle, fill=blue!20] \tikzstyle{edge} =
        [draw,thick,->] \tikzstyle{weight}= [font=\small,above]
        \tikzstyle{weight1} = [font=\small,above=0.5cm,right=0.5cm]
        \tikzstyle{weight2} = [font=\small,above=0.5cm,left=0.4cm]
        \node[vertex] (a) at (0,0) {a}; \node[vertex] (b) at (-1,-1)
        {b}; \node[vertex] (c) at (1,-1) {c}; \node[vertex] (d) at
        (-1,-3) {d}; \node[vertex] (e) at (1,-3) {e}; \path[edge] (a)
        -- node [left]{0.9} (b); \path[edge] (a) -- node [right]{0.2}
        (c); \path[edge] (b) -- node [left]{0.8} (d); \path[edge] (c)
        -- node [right]{0.1} (e); \path[edge] (b) -- node
        [above=0.2cm,left=0.2cm]{0.01} (e); \path[edge] (c) -- node
        [above=0.2cm,right=0.3cm]{0.7} (d);

      \end{tikzpicture}
    \end{minipage}
\end{adjustbox}
    & 
  \begin{tabular}{|llll}
&
\begin{adjustbox}{width=1in,keepaspectratio}
\begin{minipage}[t]{1.2in}
\begin{lstlisting}[name=IntroExample1]
% Possible edges
poss_edge(a,b).
poss_edge(a,c).
poss_edge(b,d).
poss_edge(b,e).
poss_edge(c,d).
poss_edge(c,e).
\end{lstlisting}
\end{minipage}
\end{adjustbox}
    & 
\begin{adjustbox}{width=2in,keepaspectratio}
\begin{minipage}[t]{2.4in}
\begin{lstlisting}[name=IntroExample1]
% Random processes
values(r(_,_), [t,f]).
% Distributions
:- set_sw(r(a,b), [0.9,0.1]).
:- set_sw(r(a,c), [0.2,0.8]).
:- set_sw(r(b,d), [0.8,0.2]).
:- set_sw(r(b,e), [0.01,0.99]).
:- set_sw(r(c,d), [0.7,0.3]).
:- set_sw(r(c,e), [0.1,0.9]).
\end{lstlisting}
\end{minipage}
\end{adjustbox}
    & 
\begin{adjustbox}{width=1.8in,keepaspectratio}
\begin{minipage}[t]{2.2in}
\begin{lstlisting}[name=IntroExample1]
% Edge:
edge(X,Y) :- 
     poss_edge(X,Y), 
     msw(r(X,Y),t).
% Reachability on acyclic graphs
reach(X,Y) :- edge(X,Y).
reach(X,Y) :- 
    edge(X,Z), 
    reach(Z,Y).
\end{lstlisting}
\end{minipage}
\end{adjustbox}\\
  \end{tabular}
\\
(a) & (b) \\
  \end{tabular}
  \caption{Example: (a) Probabilistic Graph;  (b) Reachability over probabilistic graphs in PRISM}
  \label{fig:intro-example}
\end{figure}

An example problem with such a model is reachability over finite probabilistic
graphs, i.e., graphs in which the presence or absence of edges is determined by
a set of independent probabilistic processes.  Fig.~\ref{fig:intro-example}(a)
shows an example of a probabilistic graph, where labels on the edges denote the
probability with which that edge is present.  For instance, edge $(a,b)$ is
present with probability $0.9$ while edge $(b,e)$ is present with probability
$0.01$.  \comment{Since edges are selected independently, the probability that
  path $(a,b), (b,e)$ exists is $0.9 \times 0.01 = 0.009$.}  The logical
relationship between reachability of $e$ from $a$, and the underlying edges in
the graph cannot be expressed concisely in standard probabilistic frameworks.  A
PRISM program\footnote{This problem has a simpler encoding in ProbLog, and can
  be encoded in easily LPAD as well.  We use PRISM since it simplifies the
  description of the techniques.}  encoding this problem is shown in
Fig.~\ref{fig:intro-example}(b).

As illustrated in Fig.~\ref{fig:intro-example}(b), PRISM adds
\emph{probabilistic facts} of the form $\kw{msw}(s,i,t)$ where $s$ is a term
representing a random process called a \emph{switch}, $i$ an instance of the
switch, and $t$ its outcome\footnote{The instance number is ommitted if only a
  single instance is used in the program.}. The range of a switch is specified
by ``\texttt{value}'' declarations; and its distribution is specified by
``\texttt{set\_sw}'' declarations.  A \emph{possible world} associates an
outcome with each switch instance, and can be seen as a set of \texttt{msw}
facts external to the program.  In each possible world, the PRISM program,
together with \texttt{msw} facts defining the world is a non-probabilistic
program; the distribution over the possible worlds induces a distribution over
the models of the PRISM program.  Such a declarative \emph{distribution
  semantics} originally defined for ICL and PRISM has been defined for other PLP
languages such as LPAD and ProbLog as well.

\lstset{%
  language=Prolog,
  basicstyle=\ttfamily,
  commentstyle=\rmfamily\it\color{DarkBlue},
  columns=fullflexible,
  numbers=left, numberstyle=\tiny, stepnumber=1, numbersep=5pt,
  firstnumber=auto,
  numberfirstline=true,
  numberblanklines=true
}


\paragraph{Motivation.}
While PLPs can concisely express such problems, typical implementations
of PLP systems have limitations.  For instance, PRISM's standard
inference technique is based on enumerating explanations for
answers, treating the set of explanations as pairwise mutually
exclusive; in fact, due to this limitation the probability of
\texttt{reach(a,e)} in the above example cannot be computed in PRISM.
ProbLog, and subsequently, PITA~\cite{riguzzi2011pita}, removed these
restrictions; however, \emph{exact} inference in these systems does
not scale beyond graphs with a few hundred vertices.

Of the several powerful sampling-based techniques developed for statistical
reasoning, Markov Chain Monte Carlo (MCMC) techniques are especially suited for
inference in PLPs, as shown by~\citet{cussens2000stochastic}
and~\citet{moldovan2013mcmc}.

\paragraph{The Problem.}
PLP queries for evaluating conditional probabilities are called as
\emph{conditional queries} and denoted as $\cond{q}{e}$, where $q$ and $e$ are
ground atomic goals, called \emph{query} and \emph{evidence}, respectively.  A
conditional query $\cond{q}{e}$ denotes a suitably normalized distribution of
$q$ over all possible worlds where $e$ holds.  Existing PLP systems either
provide efficient techniques that apply to a restricted class of $q$ and $e$
(e.g., hindsight in PRISM) or do not treat evidence specially, leading to poor
performance especially when the likelihood of evidence is low.  For instance,
consider the problem of determining the probability that $d$ is reachable from
$a$, given that $e$ is reachable from $a$ (i.e.
$\cond{\mathtt{reach(a,d)}}{\mathtt{reach(a,e)}}$, over the probabilistic graph
in Fig.~\ref{fig:intro-example}(a).  Techniques such as the one proposed
by~\citet{moldovan2013mcmc} will generate a world and reject it if evidence does
not hold in the world.  Since the probability that $e$ is reachable is
$0.02882$, a large percentage of generated worlds will be inconsistent with the
evidence, and hence unusable for computing the conditional probability.

The problem of efficiently estimating the conditional probability,
even when the likelihood of evidence is low, has remained unaddressed
in the context of PLP.  We explore this problem in this paper, by
developing an \emph{Adaptive} Markov Chain Monte Carlo (AMCMC)
technique.  Following adaptive MCMC techniques in statistical
reasoning, we progressively modify the distribution from which samples
are derived; we modify the distribution so as to favor those samples
that are consistent with evidence.  The adaptive sampler reduces the
number of generated samples needed to estimate the conditional
probability to a given precision.

\paragraph{Approach Overview.}  
Our technical development starts with a MCMC technique where each
state of the Markov Chain is an \emph{assignment} of values to a set
of switch instances.  An assignment at a state corresponds to a set of
possible worlds such that the truth values of evidence and query are identical
in all the worlds in the set.  Transitions are proposed on this chain
by resampling one or more switch instances in the state and extending
the resulting assignment to another state.  A Metropolis-Hastings
\cite{hastings1970monte} 
sampler is used to accept or reject this proposal, yielding the next
state in the chain.

To this basic MCMC technique, we introduce adaptation as follows.  For
each switch instance/outcome pair, we maintain  \emph{Q-value} which is
the likelihood that an evaluation of the evidence goal using that
switch instance/outcome will succeed.  Q-values are computed by 
maintaining a sequence of switch instances and outcomes used to evaluate the evidence
goal, and propagating rewards through this sequence depending on the
success or failure of the evaluation.  The adapted distribution of a
switch instance is proportional to the original distribution weighted
by (normalized) Q-values of each outcome.

Although motivated by problems where the likelihood of evidence is
low, the technique we describe is more generally applicable, 
even to unconditional queries.
\comment{We place no restriction on the relationship between the query
  and evidence; we consider them both as logical atoms. For instance,
  with our technique we can infer
  $\cond{\mathtt{reach(a,d)}}{\mathtt{reach(a,e)}}$; note that the
  derivations of $\mathtt{reach(a,d)}$ and $\mathtt{reach(a,e)}$ do
  not have a simple relationship (e.g. sub-derivations), and the
  evidence is a complex query itself (and not just a valuation of
  random variables).}

\comment{
This form of adaption is not limited to MCMC, and can be used for
independent sampling as well, provided that the adapted 
and original distributions are identical for consistent samples
(modulo normalizing constants).  This can be assured only if, for each
switch $r$ in $W_e$, the
consistency of an answer is conditionally independent of all
predecessors of $r$ in $W_e$ (given $r$'s outcome).  We call queries
having this property as having ``Markovian evaluation structure''.
For such queries, our adaption scheme can be used for independent
samples as well, thereby offering an alternative when MCMC itself is
slow-mixing and prohibitively expensive.
}

\paragraph{Summary of Contributions.}
\begin{enumerate}
\item We define a MCMC procedure where states of the Markov chain are
  sets of possible worlds.  This procedure is largely \emph{independent of LP
  evaluation itself}, and hence can be used for approximate inference
  in probabilistic logic programs extended with tabling, constraint
  handling, 
  or other features (Section~\ref{sec:approx}).
\item We define an adaptation procedure to modify the distribution
  from which samples are drawn.  The aim of the adaption is to
  increase the likelihood that a sample will be consistent with given
  evidence (if any).  We show that the adaption satisfies the
  ``diminishing adaption'' condition and hence can be used to
  effectively to adapt an MCMC procedure (Section~\ref{sec:amcmc}).
\item For a class of queries satisfying a
  ``Markovian evaluation structure'', the adapted distribution of a
  random variable coincides with its marginal.  For such queries, we
  obtain an alternative adaption
  procedure that can be used to obtain an adaptive independent sampler
  (Section~\ref{sec:amcmc}).
\end{enumerate}
We describe the results of our preliminary experiments to evaluate the
MCMC procedure as well as the adaptation procedure in Section~\ref{sec:expt}.
 
The rest of the paper begins with a brief
overview of MCMC in Section~\ref{sec:prelim}.  A more
detailed description of related work and concluding remarks
appears 
Section~\ref{sec:disc}.

\section{Preliminaries: Markov Chain Monte Carlo Techniques}
\label{sec:prelim}
\label{sec:prelim-mcmc}
A sequence of random variables $X^{(i)}, i \geq 0 $ taking on values $x^{(i)}$ is
called a \emph{Markov chain} if $P(x^{(i)} \vert
x^{(i-1)},x^{(i-2)},$ $\ldots,x^{(1)})=P(x^{(i)} \vert x^{(i-1)})$
\cite{andrieu2003introduction}.  The values of the random variables are chosen
from a fixed set called the state space of the Markov chain.  When the
state space is finite, the one step transition probabilities between various
states are generally given as a matrix known as the \emph{transition kernel}.

Given a distribution on the values of $X^{(0)}$, the distribution on the values
of any $X^{(i)}, i > 0$ can be computed by multiplying the transition kernel $i$
times with the initial distribution.  For certain Markov chains,
 irrespective of the initial distribution on $X^{(0)}$, the distribution on
the of values $X^{(n)}$ converges as $n$ increases, to 
its \emph{limiting distribution} or \emph{stationary distribution}.  More
formally, a stationary distribution $\pi$ with respect to a Markov chain with
transition kernel $A$ satisfies the condition $\pi = \pi A$.  
Irreducible and aperiodic Markov
chains have a unique stationary distribution~\cite{andrieu2003introduction}.
In practice, a distribution $\pi$ is verified to be a stationary
distribution, if for any two states $x$ and $y$ the following 
\emph{detailed balance} condition holds.  
\[
\pi(x) A(x,y) = \pi(y) A(y,x)
\]

Given a hard to sample \emph{target} distribution, MCMC techniques solve the
problem by constructing a Markov chain whose stationary distribution is the
target distribution and drawing samples from it.  \textbf{Metropolis-Hastings}
(MH) is a popular MCMC-based sampling technique.  Given a target distribution
$\pi$ and an irreducible, aperiodic Markov chain with transition kernel $A$, the
MH sampler proposes a transition from state $x$ to $y$ according to $A(x,y)$,
but then accepts or rejects this proposal according to the acceptance
probability $min\{1,
\frac{\pi(y)A(y,x)}{\pi(x)A(x,y)}\}$~\cite{hastings1970monte}.

\comment{ Looks like we dont use Gibbs, so I'm commenting this out:
\item \textbf{Gibbs-Sampling} Gibbs sampling requires a factorized state space
  where the state $X$ itself is divided into components $\{X_{1}, \ldots,
  X_{m}\}$, and the conditional probabilities $P(X_{i}|X)$ are known. Given the
  current state $x$, Gibbs sampling proceeds by selecting one of the components
  $X_{i}$ at random and sampling it according to $P(X_{i}|x)$. All state
  transition proposals are accepted and there is no notion of acceptance
  probability as in the MH algorithm \cite{geman1984stochastic}.
\end{itemize}
}




\subsection*{Adaptive MCMC}
\label{sec:adaptivemcmc-background}
Given a target distribution from which samples need to be generated, MCMC
algorithms such as MH construct a Markov chain whose transition kernel satisfies the
detailed balance condition with respect to the target distribution.  However,
there are several transition kernels which satisfy this requirement.  The optimal
choice is not always clear. In such situations adaptive MCMC algorithms tune the
transition kernel as the chain proceeds \cite{roberts2007coupling}.

The \emph{total variation distance} between two distributions $P$ and $Q$ is
defined as $\| P - Q \| = \frac{1}{2}\sum_{x}\vert P(x) - Q(x)\vert$
\cite{levin2009markov}.  A Markov chain with transition kernel $A$ is said to be
\emph{ergodic} with respect to a target distribution $\pi$ if $lim_{n\rightarrow
  \infty}\|A^{n}(x,.)-\pi(.)\|=0$, for all $x$~\cite{roberts2007coupling}.  In
other words, the rows of the transition matrix converge to the target
distribution as it is repeatedly multiplied with itself.  It also means that
irrespective of which state we start in the long term probability of being in a
state converges to the probability of that state given by the target
distribution.  In general, adaptation of a transition kernel does not preserve
ergodicity of the Markov chain with respect to the target distribution. However,
ergodicity has been shown to be preserved under certain conditions, namely
performing adaptation at \emph{regeneration times}
\cite{gilks1998adaptive,brockwell2005identification} and \emph{diminishing
  adaptation} \cite{roberts2007coupling,roberts2009examples}.  We use the latter
condition for preserving ergodicity in our adaptation scheme.

\paragraph{\textbf{Ergodicity conditions.}}
\label{sec:ergodicity}
Given a family of transition kernels $\{P_{\Gamma_1},P_{\Gamma_2},\ldots \}$,
each having the target distribution $\pi$ as the stationary distribution,
adaptive MCMC algorithms choose the transition kernel $P_{\Gamma_{i}}$ at time
step $i$. The update rule of $P_{\Gamma_{i}}$ is specified by the adaptive
algorithm. Then ergodicity is preserved if all transition kernels have
\emph{simultaneous uniform ergodicity}, namely, 
\[
\text{For each }\varepsilon > 0,\text{ there exists }N=N(\varepsilon) \in
\mathbb{N} \text{ such that }\|P_\gamma^N(x,.)-\pi(.)\| \leq \varepsilon
\text{ for all }x \text{ and } \gamma
\]
and the following
\emph{diminishing adaptation} condition is satisfied.
\cite{roberts2007coupling}.
\begin{equation*}
\lim_{n \to \infty} \text{sup}\|P_{\Gamma_{n+1}}(x,.)-P_{\Gamma_{n}}(x,.)\|=0
\text{ in probability}
\end{equation*}


\comment{
I added a para on PRISM to intro, so this can go!

\subsection{PRISM programs}
\label{sec:prism}

PRISM programs extend traditional logic programs by allowing
probabilistic facts (in addition to traditional non-probabilistic
facts).  Probabilistic facts are represented using \emph{msw}
atoms. An atom $\kw{msw}(\id{id}, i, v)$ represents the fact that a
\emph{switch} named $\id{id}$, which represents a random process, has
outcome $v$ in trial $i$.  The set of possible outcomes, called \emph{values}
are finite, mutually exclusive and their probabilities sum to one
\cite{sato2008glimpse}.  In PRISM, distinct switches represent
independent random processes.


A PRISM program implicitly defines a joint distribution on the set of
all probabilistic facts. A sample from this distribution gives 
values for all switches in the program.  An assignment of values to
switches is called a \emph{world}.  In each world, a switch has a
unique value.  Moreover, in each world, the PRISM program corresponds
to a non-probabilistic (definite) logic program.
The set of sampled probabilistic
facts together with the rules in the PRISM program define a least
model. Therefore a distribution on the probabilistic facts induces a
distribution over least models of the program; this is known as distribution
semantics~\cite{sato1995statistical}.

\textbf{I dont know how much detail to give about PRISM here, and what would
be relevant.  We'll come back to this part after the main technical
sections are done.
}

}

\comment{ A PRISM program is
shown in example \ref{ex:concex}. This program encodes reachability in a
directed acyclic graph (DAG) whose edges are probabilistic. The \emph{values}
and \emph{set\_sw} facts specify the values of switches and the probabilities of
those values.}

\comment{
\subsection{Evaluation trees}
\label{sec:evaltree}
In this section we formalize the notion of evaluation trees, which will be used
to generate mutually exclusive samples for inference. Before we define
evaluation trees we need to define SLD trees and refutation procedure for PRISM
programs. The notion of SLD-tree \cite{lloyd1984foundations} can be extended to
PRISM programs by defining the children of a node whose selected atom is an
\emph{msw} atom. For other nodes (i.e., nodes whose selected atom is not an
\emph{msw} atom) children are defined in the traditional way to be the
resolvants for each of the clauses whose rule head matches with the selected
atom \cite{lloyd1984foundations}.

\begin{Def}
Let $\leftarrow A_{1}, \ldots, A_{n}$, (where each $A_{i}$ is a literal) be a
goal(node) in the SLD-tree of a PRISM program. Let the computation rule select
atom $A_{j}$ and $A_{j}$ is \emph{msw(P, V)}, where $P$ is a ground term. The
children of the node are given as follows:
\begin{itemize}
\item If the switch $P$ is defined (i.e., there exist \emph{values} and
  \emph{set\_sw} atoms whose random process unifies with $P$)
  \begin{itemize}
    \item If $V$ is a variable then a child $\leftarrow (A_{1}, \ldots,
      A_{j-1}, A_{j+1}, \ldots, A_{n})\theta$ is given for each substitution
      $\theta = \{V/v\}$ where $v$ is a value of the switch $P$.
    \item If $V$ is a value then one child is $\leftarrow A_{1},A_{2}, \ldots,
      A_{j-1}, A_{j+1}, \ldots, A_{n}$. Other children are the leaves $\leftarrow
      fail_{v}$, where $v$ is a value of the switch $P$.
  \end{itemize}
\item If the switch $P$ is not defined, then the node is a leaf.
\end{itemize}
\end{Def}

The PROLOG refutation procedure \cite{lloyd1984foundations} is also extended to
PRISM case, by adapting it to deal with \emph{msw} atoms. When exploring
goals(nodes) which don't have \emph{msw} atom as selected atom (i.e., leftmost
atom), the refutation procedure is same as the PROLOG refutation procedure
(i.e., depth-first search for success branch with clauses being considered in
the textual order). The search strategy when exploring goals with \emph{msw}
atom as the leftmost atom is defined below.

\begin{Def}
The refutation procedure starts with an empty sequence called \emph{state}. When
it encounters a goal $\leftarrow msw(P, V), A_{1}, \ldots, A_{n}$ it proceeds as
follows:
\begin{itemize}
\item If the \emph{state} contains an ordered pair $(P, O)$, then:
  \begin{itemize}
    \item If $V$ is a variable, then go to the child given by $\leftarrow
      (A_{1}, \ldots, A_{n})\theta$ where $\theta=\{V/O\}$.
    \item Else if $V=O$ then go to the child $\leftarrow A_{1},
      \ldots, A_{n}$, else go to the child $fail_{O}$.
  \end{itemize}
\item Else
  \begin{itemize}
    \item If $V$ is a variable, then choose one of the children $\leftarrow
      (A_{1}, \ldots, A_{n})\theta$ where $\theta=\{V/v\}$ with probability of
      value $v$ and append the ordered pair $(P, v)$ to \emph{state}.
    \item Else choose one of the children of the node according to the
      probability of the value $V$ and the probabilities of $v$ for various
      $fail_{v}$ nodes. Add the ordered pair $(P, V)$ or $(P, v)$ based on the
      node chosen.
  \end{itemize}
\item Prune the subtrees of all children except the one chosen.
\end{itemize}
\end{Def}

We restrict our attention to PRISM programs and ground queries on PRISM programs
which have finite SLD-trees. Given that we have a finite SLD-tree there will be
finitely many \emph{states} of refutation procedure when it has succeeded or
failed. We can further show that all these \emph{states} share non-empty
prefixes, in particular all of them have the same switch for the first ordered
pair in the sequence. These states are arranged as an evaluation tree as
follows.

\begin{Def}
For each \emph{state} $\{(P_{i}, V_{i})\}_{i=1}^{n}, n \geq 1$, of the
refutation procedure which ended in success or failure nodes and edges are added
to the evaluation tree as follows:
\begin{itemize}
  \item If the evaluation tree is empty add root with the label $P_{1}$.
  \item For i = 1 to n
  \begin{itemize}
    \item Starting from root trace the path from $P_{1}$ to $P_{i}$ by following
      edges labeled $V_{1}$ to $V_{i-1}$. If any nodes or edges are non-existent
      add them.
  \end{itemize}
\end{itemize}
\end{Def}

\begin{Ex}
Consider the program shown in figure \ref{code:reach}. It encodes reachability
in a DAG whose edges are probabilistic (i.e., exist with probability 0.5). The
graph encoded is a four node DAG and we wish to compute the probability of
reaching one node from another. The SLD-tree for the query $reach(2, 4)$ is
given in figure \ref{fig:reachsld}. The refutation procedure can take several
alternate paths through the SLD-tree. The evaluation tree for the same query is
shown in figure \ref{fig:reacheval}.
\label{ex:concex} 
\end{Ex}


\begin{figure}
\hrulefill
\begin{center}
\begin{verbatim}
reach(X, X).
reach(X, Y) :- pedge(X, Y).
reach(X, Y) :- pedge(X, Z), reach(Z, Y).
pedge(X, Y) :- edge(X, Y), msw(e(X, Y), t).
edge(1, 2). edge(1, 4). 
edge(2, 4). edge(2, 3). 
edge(3, 4).
values(_, [t, f]).
set_sw(_, [0.5, 0.5]).
\end{verbatim}
\end{center}
\caption{Graph reachability program}
\hrulefill
\label{code:reach}
\end{figure}

\begin{figure}
\begin{tikzpicture}
\tikzset{level 1/.style={sibling distance=60mm}}
\tikzset{level 3/.style={sibling distance=50mm}}
\tikzset{level 4/.style={sibling distance=15mm}}
\tikzset{level 5/.style={sibling distance=50mm}}
\tikzset{level 8/.style={sibling distance=20mm}}
\node (r) {reach(2,4)}
child {
  node (l11) {pedge(2,4)}
  child {
    node (l21) {edge(2,4), msw(e(2,4), t)}
    child {
      node (l31) {msw(e(2,4), t)}
      child {
        node (l41) {$\proofbox$}
      }
      child {
        node (l42) {$\text{fail}_{\text{f}}$}
      }
    }
  }
}
child {
  node (l12) {pedge(2,Z), reach(Z,4)}
  child {
    node (l22) {edge(2,Z), msw(e(2,Z), t), reach(Z,4)}
    child {
      node (l32) {msw(e(2,3), t), reach(3,4)} 
      child {
        node (l43) {reach(3,4)}
        child {
          node (l51) {pedge(3,4)}
          child {
            node (l61) {edge(3,4), msw(e(3,4), t)}
            child {
              node (l71) {msw(e(3,4), t)}
              child {
                node (l81) {$\proofbox$}
              }
              child {
                node (l82) {$\text{fail}_{\text{f}}$}
              }
            }
          }
        }
        child {
          node (l52) {pedge(3,Z), reach(Z,4)}
          child {
            node (l62) {edge(3,Z), msw(e(3,Z), t), reach(Z,4)}
            child {
              node (l72) {msw(e(3,4), t), reach(4,4)}
              child {
                node (l83) {reach(4,4)}
                child {
                  node (l91) {$\proofbox$}
                }
              }
              child {
                node (l84) {$\text{fail}_{\text{f}}$}
              }
            }
          }
        }
      }
      child {
        node (l44) {$\text{fail}_{\text{f}}$}
      }
    }
    child {
      node (l33) {msw(e(2,4), t), reach(4,4)}
      child {
        node (l45) {reach(4,4)}
        child {
          node (l53) {$\proofbox$}
        }
      }
      child {
        node (l46) {$\text{fail}_{\text{f}}$}
      }
    }
  }
};
\end{tikzpicture}
\caption{SLD-tree for the query reach(2, 4)}
\label{fig:reachsld}
\end{figure}


\begin{figure}
\begin{tikzpicture}
\node (r) {msw(e(2,4))}
child{
  node (l11) {$\proofbox$} edge from parent node[left] {t}
}
child{
  node (l12) {msw(e(2,3))} 
  child{
    node (l21) {msw(e(3,4))}
    child {
      node (l31) {$\proofbox$} edge from parent node[left] {t}
    }
    child {
      node (l32) {fail} edge from parent node[right] {f}
    } edge from parent node[left] {t}
  }
  child {
    node (l22) {fail} edge from parent node[right] {f}
  } edge from parent node[right] {f}
};
\end{tikzpicture}
\caption{Evaluation tree for reach(2,4)}
\label{fig:reacheval}
\end{figure}

}

\section{MCMC for Probabilistic Logic Programs}
\label{sec:approx}

The ability to treat a PRISM program as non-probabilistic in each
world also helps us in designing sample-based query evaluation.  Given
a PRISM program $P$ and a ground goal $q$, we lazily construct a set
of worlds by sampling, such that $q$ succeeds or fails in all worlds
in the set.  The set of worlds are represented by \emph{assignments}
described below.

\subsection{Sample-Based Query Evaluation}
\paragraph{\textbf{Assignments.}}
We use a structure called an 
\emph{assignment} to 
 keep track of known outcomes of switch instances when sampling from a
 PRISM program.
An \emph{assignment}  is denoted by partial function $\sigma$ such that
$\sigma(s, i)$ is the value of instance $i$ of switch $s$.  Note that
$\sigma$ represents a set of worlds; the set of worlds corresponding
to $\sigma$ is denoted by $\id{worlds}(\sigma)$.

Let $\sigma$ be an assignment.  We say $\sigma' =
\sigma[(s,i)\rightarrow v]$ is an assignment that is identical to
$\sigma$ at every point except at $(s,i)$ where $\sigma'(s,i) =
v$.  We define a partial order ``$\succeq$'' over assignments:
$\sigma' \succeq \sigma$ if $\sigma'(s,i) = \sigma(s,i)$ whenever
$\sigma(s,i)\not=\bot$.  We also say that $\sigma'$ \emph{extends}
$\sigma$ if $\sigma' \succeq \sigma$.  
We say that two assignments $\sigma$ and $\sigma'$ are \emph{mutually
exclusive}, denoted by $\sigma \parallel \sigma'$,  if there is some switch instance $(s,i)$ such that
both $\sigma$ and $\sigma'$ are defined at $(s,i)$, but
$\sigma(s,i) \not = \sigma'(s,i)$.  Two assignments are
\emph{compatible} (denoted
by ``$\nparallel$'') if they are 
\emph{not} mutually exclusive.

Given a switch $s$, we denote by $\id{random}(s)$ a value
randomly drawn from the domain of $s$ using the probability
distribution defined for $s$.  For looking up in an assignment or
extending an assignment, we define a function \id{pick\_value},
defined as follows:
\[
\begin{array}{rcl}
  \id{pick\_value}(\sigma, s, i) & =  &
  \left\{
    \begin{array}{ll}
      \langle v, \sigma\rangle & \mbox{if $\sigma(s,i) = v \not= \bot$ }\\
      \langle v, \sigma[(s,i)\mapsto v]\rangle & \mbox{if $\sigma(s,i) =
        \bot$ and $v = \id{random}(s)$}\\
    \end{array}
  \right.
\end{array}
\]
Note that $\id{pick\_value}$ is non-decreasing in the sense that if
$\langle v, \sigma'\rangle = \id{pick\_value}(\sigma, s,i)$, then
  $\sigma' \succeq \sigma$.  
Alternatively, we can view $\id{pick\_value}(\sigma, s, i)$ as defining a
distribution, 
generating $\langle v, \sigma\rangle$ with probability
$1$  if $\sigma(s,i)=v$; and 
$\langle v, \sigma'\rangle$ where $\sigma'=\sigma[(s,i)\mapsto v]$ with probability
$P_s(v)$  if $\sigma(s,i)=\bot$.

\paragraph{\textbf{Sampling Evaluators.}}
We describe the MCMC algorithm parameterized with respect to
a probabilistic query evaluation procedure called a \emph{Sampling
  Evaluator}.  
This permits us to describe a generic MCMC algorithm that can be
instantiated to extended probabilistic logic programming systems,
including tabled and/or constraint probabilistic LPs.

A sampling evaluator, given an assignment $\sigma$ and ground goal
$q$, (probabilistically) generates an answer to $q$ (\id{success}/\id{failure}), denoted by
$\id{ans}(q)$, an assignment $\sigma'$, and a sequence $\rho$ of
switch/instance/outcome triples $(s_1,i_1,v_1), (s_2, i_2,v_2), \ldots,
(s_k, i_k,v_k)$, $k\geq 0$
 such that the following conditions hold:
 \begin{description}
 \item[SE1.] Consider the sequence of assignments $\sigma_0, \sigma_1, \ldots, \sigma_k$ such
   that $\sigma_0 = \emptyset$, and $\sigma_{j+1} = \sigma_j[(s_{j+1},
   i_{j+1}) \mapsto v_{j+1}]$ for $0 \leq j < k$.  Then, $\sigma_k =
   \sigma'$.  Moreover, $\sigma'$ is compatible with $\sigma$, i.e.
   $\sigma' \nparallel \sigma$.
 \item[SE2.] If $\id{ans}(q) = \id{success}$ (similarly, $\id{failure}$),
   then $q$ is \emph{true} (or \emph{false}, resp.) in all worlds
   $w \in \id{worlds}(\sigma')$.
\end{description}
Moreover, let $\Sigma$ denote the set of all $\sigma'$
generated by the sampling evaluator.  Then,
\begin{description}
\item[SE3.] If  $w$ is a world s.t. $q$ is true (similarly, false) in $w$, then 
  $\exists \sigma' \in \Sigma$ such that $w \in \id{worlds}(\sigma')$ and
  $\id{ans}(q) = \id{success}$ (or $\id{failure}$, respectively).
\item[SE4.] Every distinct $\sigma_a, \sigma_b \in \Sigma$ are
  mutually exclusive: i.e. either $\sigma_a=\sigma_b$, or
  $\sigma_a \parallel \sigma_b$.
\end{description}

Properties \textbf{SE2} and \textbf{SE3} correspond to soundness and
completeness, respectively.  Property \textbf{SE4} ensures that a
sampling evaluator can be used to refine states in an MCMC algorithm.
Not all evaluators in a probabilistic logic programming system may
satisfy \textbf{SE4}.  For instance, if an evaluator is based on
constructing explanations for a goal (e.g. as in Problog) since two distinct explanations may not
be mutually exclusive, it will violate \textbf{SE4}.  \citet{moldovan2013mcmc} overcome this
problem by using the Karp-Luby algorithm, resampling switch instances
to eliminate overlaps in states.  In contrast, we show below that we can use
Prolog-style evaluation, performed  until the first derivation
is found (if one exists) to construct a sampling evaluator
i.e. one satisfying the above requirements including \textbf{SE4}.

\paragraph{\textbf{A sampling evaluator for non-tabled PRISM.}}
Figure~\ref{fig:sampler} shows the sampling evaluator for non-tabled
PRISM programs, constructed by extending the well-known Prolog
meta-interpreter.  The assignment and the switch/instance/outcome
sequence are maintained in the dynamic database.  Observe that the
evaluation follows that of Prolog as long as the selected literal is
not an \texttt{msw}.  When the selected literal is of the form
$\kw{msw}(s, i, t)$, we get the value of $(s,i)$ using
$\id{pick\_value}$, and record this selection.  When evaluation
produces the empty clause, we know that $q$ has a derivation in all
worlds consistent $\sigma$.  When every proof finitely fails, we know
that $q$ has no derivation in any world that is consistent with
$\sigma$.  Thus the evaluator in Fig.~\ref{fig:sampler} has the
properties described for a sampling evaluator.  To see how the
procedure satisfies condition \textbf{SE4}, assume to the contrary
that the procedure has two executions that generate two assignments,
$\sigma_a$ and $\sigma_b$.  Note that the procedure is deterministic
and takes the same sequence of steps, except when an \texttt{msw} is
encountered.  If the two executions picked different outcomes, then
$\sigma_a \parallel \sigma_b$.  Otherwise, we can show by induction
on the sizes of $\sigma_a$ and $\sigma_b$, that the assignments are
either identical or mutually exclusive.

\begin{figure}
  \begin{adjustbox}{width=6.5in,keepaspectratio}
    \begin{tabular}{ll}
    \begin{minipage}[t]{3.5in}
\begin{lstlisting}[name=Sampler]
% Given assignment is in sigma/3;  
% computed assignment is in sigma_prime/3.
% rho/3 is sequence of switch/instance/outcomes. 
% sigma_prime/3 and rho/3 are initially empty
:- dynamic sigma/3, sigma_prime/3, rho/3.

% Sampling Evaluator for a ground goal G:
sample_eval(G) :- eval(G), !.

eval(true) :- !.
eval((G1,G2)) :- !, eval(G1), eval(G2).
eval((G1;G2)) :- !, eval(G1); eval(G2).
eval(msw(S,I,V)) :- !, pick_value(S,I,V).
eval(G) :- clause(G, B), eval(B).
\end{lstlisting}
\end{minipage}
&
    \begin{minipage}[t]{3.5in}
\begin{lstlisting}[name=Sampler]
% Pick value from sigma,
% extending it via sampling if necessary.
pick_value(S,I,V) :-
    (sigma_prime(S,I,U)  % if already is defined
     -> true
      ; (sigma(S,I,U)    % if in current assignment
        -> assert(sigma_prime(S,I,U)),
         % genrandom generates a random value U 
         % according to the distribution of S
         ; genrandom(S,U), 
           assert(sigma_prime(S,I,U))
    ), assertz(rho(S,I,U)),  % update sequence
    V=U.   % ensure sigma_prime and rho are
           % updated regardless of given V
\end{lstlisting}
\end{minipage}
\end{tabular}
\end{adjustbox}
\caption{Sampling evaluator derived from Prolog meta-interpreter}
\label{fig:sampler}
\end{figure}

\subsection{MCMC-Based Inference of Conditional Probabilities}
\paragraph{\textbf{Initial Sample.}}
To perform inference using MCMC, we need (1) a way to randomly
generate an initial state, and (2) a way to generate the successor
state of a given state.  The Markov Chain we construct has
assignments as states.  When evaluating probabilities of unconditional
queries, we can generate an assignment corresponding to the initial
state by invoking a sampling evaluator with an empty assignment.  For
conditional queries, we construct a Markov chain whose initial state
as well as other states considered in a run are all consistent with
evidence, as follows.

\comment{
Consider evaluating conditional probability $\cond{q}{e}$.  We can
sample a run of the Markov Chain, counting the states where evidence
$e$ holds as $N$, and counting the states where both the evidence and
query hold as $N_q$ .  The estimate of $\cond{q}{e}$ is $N_q/N$.  Note
that visiting states where evidence does not hold is wasted effort
(since they do not contribute directly to the probability estimate).
Hence we construct a Markov Chain where the evidence holds in every
state.
}

A randomly constructed explanation for evidence is used to
generate the initial state.  We do this via a Prolog-style
backtracking search for a derivation of evidence, and collect all the
switches and valuations used in that derivation into an initial
assignment.  To ensure that the initial assignment is randomly
selected, we randomize the order in which clauses and switch values
are selected during the backtracking search.  We
refer to this procedure as $\id{InitialSample}(P,e)$ 
in the MCMC-based algorithm for inferring
conditional probabilities shown in Fig.~\ref{alg:mcmc}.

\begin{figure}
  \centering
  \begin{adjustbox}{width=3.5in,keepaspectratio}
    \begin{minipage}{3.8in}
\begin{algorithmic}[1]
\Function{MCMC}{}
   \State \textbf{Input:} $P$: Program, $q$: Query, $e$: Evidence,
    \State  $\quad\quad$  $N$: Steps to simulate
    \State \textbf{Output:} $p = \cond{q}{e}$
    \Statex \textcolor{DarkBlue}{\quad\ //  Initialize }
    \State $\sigma_0$ := $\id{InitialSample}(P, e)$
    \State $(r_q, \sigma, \_)$ := $\id{SamplingEvaluator}(P, q, \sigma_0)$
    \State $N_q$ := $0$
    \Statex \textcolor{DarkBlue}{\quad\ //  Generate a chain of length $N$}
    \For{$N$ times}
        \State $\sigma'$ = $\id{Resample}(\sigma)$
        \State $(r'_e, \sigma_e, \rho) = \id{SamplingEvaluator}(P, e, \sigma')$
        \If{$r'_e$ = \id{success}}
        \State $(r'_q, \sigma_q, \_) = \id{SamplingEvaluator}(P, q, \sigma_e)$
        \If{$\id{accept}(\sigma, \sigma_q)$}
        \State$\sigma$ := $\sigma_q$
        \State $r_q := r'_q$
        \EndIf
        \EndIf
        \If{$r_q$ = \id{success}}
        $N_q$ := $N_q+1$
        \EndIf
        \EndFor
        \State \Return $N_q/N$
\EndFunction
\end{algorithmic}
    \end{minipage}
  \end{adjustbox}
  \caption{MCMC Algorithm for Inferring Conditional Probabilities}
\label{alg:mcmc}

\end{figure}

\paragraph{\textbf{Transitions.}}
Consider a state in the Markov Chain corresponding to assignment
$\sigma_j$.  We can generate a successor state by (1) generating an
alternative assignment $\sigma'$ by assigning different outcomes for
some switch instances in $\sigma_j$, and (2) invoking
\id{SamplingEvaluator} with $\sigma'$ to evaluate $e$ and $q$ to
obtain the next state $\sigma_{j+1}$.  The switch instances to 
be resampled can be selected in several ways.  We use one of
the two following schemes:

\medskip
\noindent
\textbf{1. Single Switch:}  We select a single $(s,i)$ such that
  $\sigma_j(s,i) \neq \bot$ uniformly, and generate $\sigma' =
  \sigma_j[(s,i) \mapsto \bot]$, effectively forgetting $(s,i)$.

\medskip
\noindent
\textbf{2. Multi-Switch:}  This resampling mode is parameterized
  with a probability $P$.  We generate $\sigma'$ from $\sigma_j$ by
  forgetting with probability $P$ each $(s,i)$ for which $\sigma_j$ is defined.

In Fig.~\ref{alg:mcmc}, the resampling procedure is referred to as
$\id{Resample}$.

\paragraph{\textbf{Metropolis-Hastings.}}
The target distribution from which we want to draw samples is 
$\cond{q}{e}$, which is proportional to $\id{prob}(q)$ when we
consider only states where $e$ holds. The proposal distribution is the
stationary distribution of a Markov Chain constructed by choosing an
initial state and making transitions as described above.  In order to
draw samples the target distribution, we construct an MH sampler as
follows.  If the proposed state is inconsistent with evidence, it is
rejected deterministically.  If it is a consistent, it is
accepted/rejected based on \emph{acceptance probability}.

For single switch resampling strategy, the acceptance probability to
go to state $\sigma_2$ from $\sigma_1$ is $\min\{1, \frac{\vert \sigma_1
  \vert}{\vert \sigma_2 \vert}\}$.   For multi-switch resampling
strategy, the acceptance probability is $1$.
The derivations of these probabilities is shown in
Appendix.

\comment{
The rate at which samples are rejected
deterministically based on the evidence (due to failure of condition
in line~11 of Fig.~\ref{alg:mcmc}) is called the \emph{rejection rate}.
Improvements can be made by using
the rejected samples to adapt the proposal distribution. This is
described in the next section.
}

\section{Adaptive MCMC for Probabilistic Logic Programs}
\label{sec:amcmc}
The rate at which samples are rejected deterministically based on the
evidence (due to failure of condition in line~11 of
Fig.~\ref{alg:mcmc}) is called the \emph{rejection rate}.  In this
section, we present a technique to progressively adapt the proposal
distribution based on the samples that have been generated so far.
The basic idea behind the adaptation scheme is that samples drawn in
the past give information about whether or not outcomes of switch
instances lead to consistent samples.

The adaptation algorithm we present here inspired by Q-learning, a reinforcement
learning technique~\cite{sutton1998introduction}.  For each distinct switch
instance/outcome triple used by the sampling evaluator, we maintain a real
number in $[0,1]$, called its \emph{Q-value}.  Intuitively the Q-value of
instance $i$ of switch $s$ for outcome $v$, denoted by $Q(s,i,v)$ represents the
probability of generating a consistent sample, when the sampling evaluator
chooses $v$ as the outcome when $(s,i)$'s value is picked using the
\emph{pick\_value} function.

\begin{figure}
\centering

\vspace*{6pt}
  \begin{adjustbox}{width=3in,keepaspectratio}
    \begin{minipage}{3in}
\begin{algorithmic}[1]
\Function{Adapt}{}
\State \textbf{Input:} $\rho$, $r$: Reward
\State \textbf{Global:} $Q$: Q-values, $c$: counts,
\State \quad\quad \quad\quad$t$: total Q-values.
\State $j$ := $\id{length}(\rho)$ \Comment{Initialize}
\While{$j > 0$}
\State \textbf{let} $(s_j, i_j, v_j) = \rho[j]$
\State $t(s_{j}, i_{j}, v_{j}) := (t(s_{j}, i_{j}, v_{j}) + r)$
\State $c(s_{j}, i_{j}, v_{j}) := c(s_j, i_j, v_j)+1$ 
\State $Q(s_{j}, i_{j}, v_{j}) := t(s_j, i_j, v_j) \div  c(s_j, i_j, v_j)$ 
\State $j := j - 1$
\State $r := \displaystyle \sum_{v \in values(s_{j})} P(s_{j}, i_{j}, v) * Q(s_{j}, i_{j},
  v)$
\EndWhile
\EndFunction
\end{algorithmic}
\end{minipage}
\end{adjustbox}
\caption{Adaptation of Q-values}

\label{alg:adapt}
\end{figure}

Initially, all Q-values are set to $1$,
representing the belief that all outcomes are equally likely to yield
consistent samples.  At each iteration of MCMC, 
adaptation is done after evidence is evaluated, by passing rewards to
each switch/instance/outcome triple in $\rho$
(computed in line~10 of Fig~\ref{alg:mcmc}).  
We begin this processing with $\id{reward}=0$  if $r'_e = \id{failure}$,
denoting an inconsistent sample, and $\id{reward}=1$ otherwise.  
We work backwards through
the sequence $\rho$ so that the last switch/instance/outcome is given a reward
of 0/1, which it then modifies and passes to the switch/instance/outcome
preceding it in $\rho$. The Q-value of each random process/instance/outcome is
computed as the average of the all rewards received by it. The algorithm for
maintaining Q-values is given in Fig.~\ref{alg:adapt}.




The MCMC algorithm in Fig.~\ref{alg:mcmc} is modified for
adaptive sampling as follows.  First of all, function \emph{\textsc{Adapt}}
is invoked after line~16.  Secondly, $pick\_value$
function used in the sampling evaluator 
draws values for a switch instance $(s,i)$ based on 
the normalized product of the original
distribution and the Q-values of $(s,i)$.  Finally, 
the acceptance probability computation is modified to take the adapted
distributions into account.  Consider computing the acceptance
probability to transition from state $\sigma$ to $\sigma'$.  We can
partition the assignment $\sigma$ into three non-overlapping
functions: $\sigma_1$ for those $(s,i)$'s defined by $\sigma$ but not
by $\sigma'$;  $\sigma_2$ for those defined by both $\sigma$ and
$\sigma'$ but assigned different values; and finally,   $\sigma_3$ for
those defined by both $\sigma$ and
$\sigma'$ and assigned same values.  We can similarly partition
$\sigma'$ into $\sigma'_1$, $\sigma'_2$ and $\sigma'_3$.

For single-switch resampling strategy, the
acceptance probability is given by
\[
\min(1, \frac{P(\sigma'_{1})P(\sigma'_{2})P'(\sigma_{1})P'(\sigma_{2})1/\vert \sigma'
  \vert}{P(\sigma_{1})P(\sigma_{2})P'(\sigma'_{1})P'(\sigma'_{2})1/\vert
  \sigma \vert})
\]
where $P$ is the original probability and $P'$ is the adapted
probability.
For multi-switch strategy, the acceptance probability is given by
\[
\min(1, \frac{P(\sigma'_{1})P(\sigma'_{2})P'(\sigma_{1})P'(\sigma_{2})}{P(\sigma_{1})P(\sigma_{2})P'(\sigma'_{1})P'(\sigma'_{2})})
\]
The derivations of these probabilities are shown in the
appendix.

\begin{Thm}
The adaptive MCMC algorithm preserves ergodicity with respect to the target
distribution.
\begin{proof}
In order to prove the theorem we need to establish the two conditions of
ergodicity mentioned in Section \ref{sec:ergodicity}. We first prove the
diminishing adaptation condition. Consider any switch/instance/value $(s,i,v)$
that receices a new reward of $\alpha$. Then the difference between successive
Q-values is
\[
\frac{c(s,i,v)*Q(s,i,v) + \alpha}{c(s,i,v)+1} - Q(s,i,v) = \frac{\alpha -
Q(s,i,v)}{c(s,i,v)+1}. 
\]
We know that $\vert \alpha - Q(s,i,v) \vert \leq 1$. Hence, as $c(s,i,v)$
increases $Q(s,i,v)$ converges.Next, we use corollary 3 and Lemma 1 of
\cite{roberts2007coupling} to prove simultaneous uniform ergodicity. As
described in the paper, fix $\varepsilon > 0$. Define $\mathcal{W}_{n} \subseteq
\mathcal{X} \times \mathcal{Y}$ to be the set of all pairs $(x,\gamma)$ such
that $\|P_{\gamma}^{n}(x,.)-\pi(.)\| < \varepsilon$, for each $n \in
\mathbb{N}$. Consider the topology defined by $\emptyset \bigcup_{n}
\mathcal{W}_{n}$ on $\mathcal{X} \times \mathcal{Y}$. It is easy to see that
$\mathcal{X} \times \mathcal{Y}$ is compact in this topology. Now consider the
distance metrics $dist(x, x') = c, c > 0$ and $dist(\gamma, \gamma') =
sup_{x,y}\|P_{\gamma}(x,.) - P_{\gamma'}(y,.)\|$ on $\mathcal{X}$ and $\mathcal{Y}$ respectively.
Given $x \in \mathcal{X}, \gamma \in \mathcal{Y}, \varepsilon > 0$, we define
$\delta(x,\gamma,\varepsilon) = c+\varepsilon$. Given this definition of
$\delta$ we can see that for all $x' \in \mathcal{X}$ and $\gamma' \in
\mathcal{Y}$ such that $dist(x,x')+dist(\gamma,\gamma') < \delta$,
$\|P_{\gamma'}(x', .) - P_{\gamma}(x, .)\| < \varepsilon$. Therefore by
corollary 3 of \cite{roberts2007coupling} we can say that our adaptive mcmc
algorithm preserves ergodicity with respect to the target distribution.

\end{proof}
\end{Thm}

\paragraph{\textbf{Beyond MCMC.}} It should be noted that the adapted
distribution need not coincide with the conditional distribution
$\cond{q}{e}$. This precludes the use of adaptation for other sampling
strategies such as independent sampling. This is not a problem for
MCMC, since the adapted distribution is used as the proposal. 
However, for a class of program/query pairs whose sampling evaluations
is ``Markovian'', each switch instance's adapted distribution
converges to its marginal distribution.  For such programs, a modified
adaptation can be used for independent sampling as well.

Consider the class of programs and queries for which the sequence $\rho$ of
random process/instance outcomes
$(s_{1},i_{1},v_{i}),\ldots,(s_{k},i_{k},v_{k})$ is such that the probability
$P(ans(e) | (s_{j},i_{j},v_{j}))$ is independent the triples
$(s_{l},i_{l},v_{l}), l < j$.  These program/query pairs are said to have a
\emph{Markovian Evaluation Structure}.  Instead of defining the Q-value to be
the average of rewards, we redefine it to be the last reward.  It can be
shown that the rewards received by any switch instances will be monotonically
decreasing during the execution of the algorithm.  This allows us to perform
independent sampling as well as MCMC for such programs and queries.

\section{Experimental Results}
\label{sec:expt}

The MCMC algorithm was implemented in the XSB logic programming
system~\cite{XSB}.  The sampling evaluator (Fig.~\ref{fig:sampler})
and the main control loop (Fig.~\ref{alg:mcmc}) were implemented in
Prolog.  Lower level primitives managing the maintenance of
assignments, resampling, computation of acceptance/rejection were
implemented in C and invoked from Prolog.
We evaluated the performance of this implementation on four synthetic
examples: BN, Hamming, Grammar, and Reach.
The experiments were run on a machine with 2.4GHz Opteron 280
processor and 4G RAM.

\begin{figure}
\begin{center}
\begin{tabular}{p{3in}p{3in}}
  \begin{tabular}[t]{l}
\includegraphics[width=2.75in]{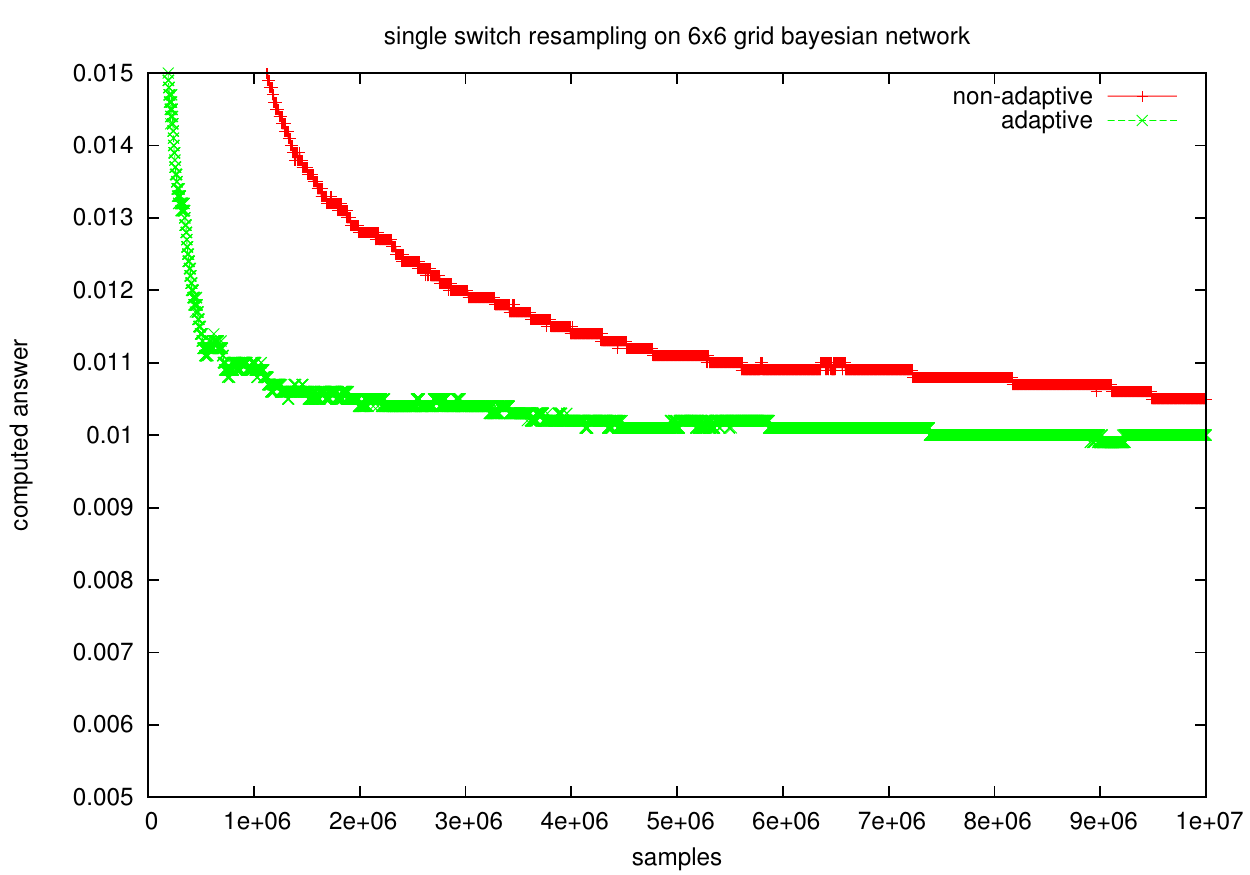}
\end{tabular}
&
  \begin{tabular}[t]{l}
\includegraphics[width=2.75in]{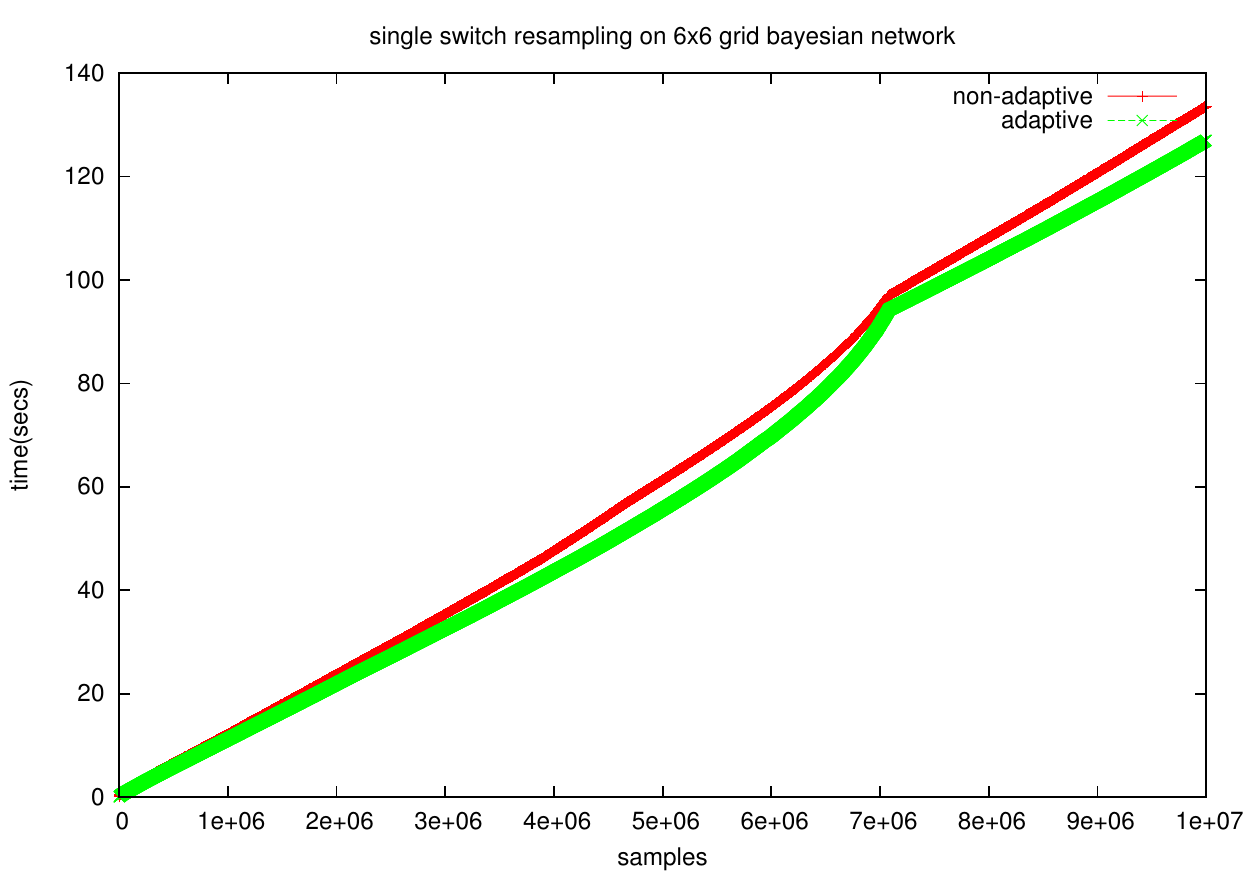}
\end{tabular}\\
(a) & (b)\\
\end{tabular}
\end{center}
\caption{Bayesian Network, $6\times6$ grid.  (a) Conditional probability computed
and (b) Running times as functions of sample size.}
\label{fig:gridtime}
\end{figure}

\paragraph{\textbf{BN.}} 
This example consists of Bayesian networks whose Boolean-valued variables  are 
arranged in the form of a grid $6\times6$ with each node having its left and top
neighbors (if any) as parents.  Evidence sets the outcome of $6$
variables; and we query the outcome of one of
the remaining variables.  Fig~\ref{fig:gridtime} shows the
conditional probability estimated by our 
algorithm, and the time taken,  both plotted as functions of sample size.
Observe from the figure that the time overhead for performing
adaptation is small.  This example clearly illustrates 
the benefit of adaptation.

\begin{figure}
\begin{center}
\begin{tabular}{p{3in}p{3in}}
  \begin{tabular}[t]{l}
\includegraphics[width=2.75in]{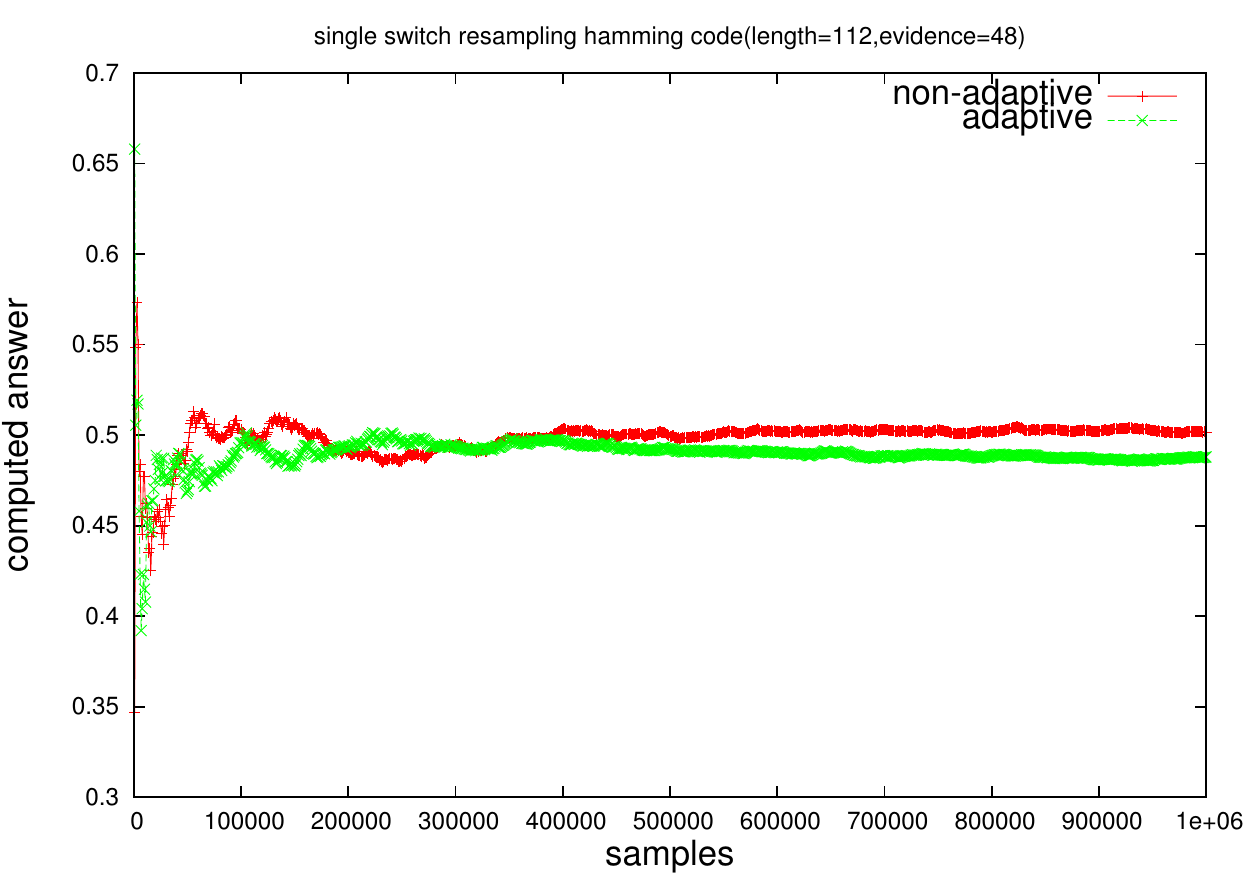}
\end{tabular}
&
  \begin{tabular}[t]{l}
\includegraphics[width=2.75in]{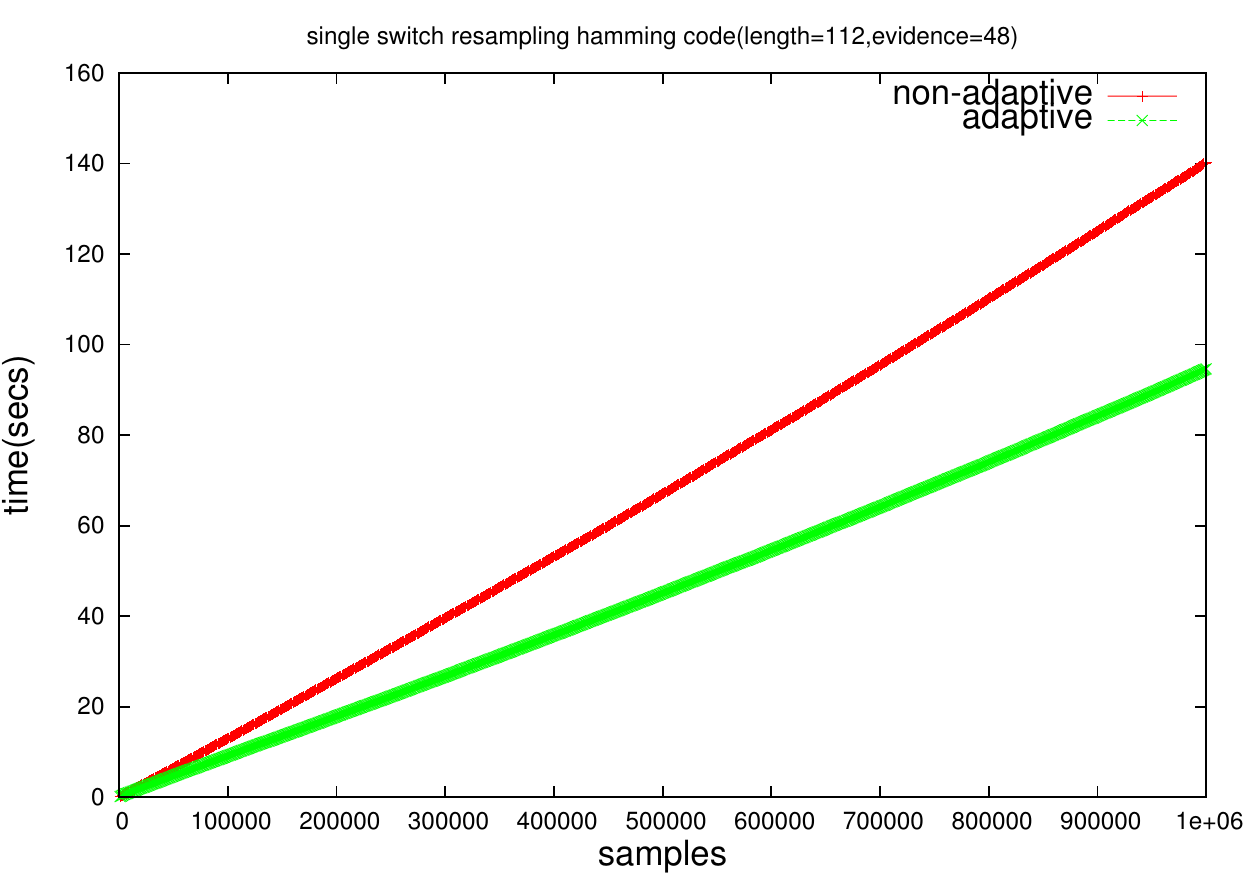}
\end{tabular}\\
(a) & (b)\\
\end{tabular}
\end{center}
\caption{Hamming Code, 112 bits.  (a) Condition probability computed
and (b) Running times as functions of sample size.}
\label{fig:hammingtime}
\end{figure}

\paragraph{\textbf{Hamming.}}
The Hamming code example is a PRISM program that generates a set of
(4,3) Hamming codes.  The evidence is a set of bits in the code with
fixed values, and the query is the value of a non-evidence
bit~\cite{moldovan2013mcmc}.  The data bits in the code were
independent random variables, while the parity bits were computed from
the data bits' values.  The answers computed by adaptive and non-adaptive samplers are
given in Fig~\ref{fig:hammingtime}(a).  
The time taken by both the samplers is shown in
Fig~\ref{fig:hammingtime}(b).  In this example, the convergence of the
adaptive MCMC is only a little better than that of the non-adaptive
algorithm.

\paragraph{\textbf{Grammar.}}
This example checks a property of strings over open and closed parentheses.
 For any randomly generated string (not necessarily balanced), we define a ``maximum nesting
level'' as the largest number of unmatched open parenthesis in a
left-to-right scan of the string.  Given that a randomly generated
string has balanced parenthesis, this example evaluates the
conditional probability of the query that determines whether a given random
string has maximum nesting level of 3 or more.  For the experiment, we
fixed the length of strings to 200.  The answers computed by 
adaptive and non-adaptive sampler are shown in
Fig~\ref{fig:grammaranswer}(a) and the times taken are shown in
Fig~\ref{fig:grammaranswer}(b).  We used
multi-switch resampling since the entire state space is not accessible
via single switch resampling.  Observe that adaptive sampling
converges faster, although by a small margin; but adaptive sampling is
almost twice as slow per iteration.

\begin{figure}
\begin{center}
\begin{tabular}{p{3in}p{3in}}
  \begin{tabular}[t]{l}
\includegraphics[width=2.75in]{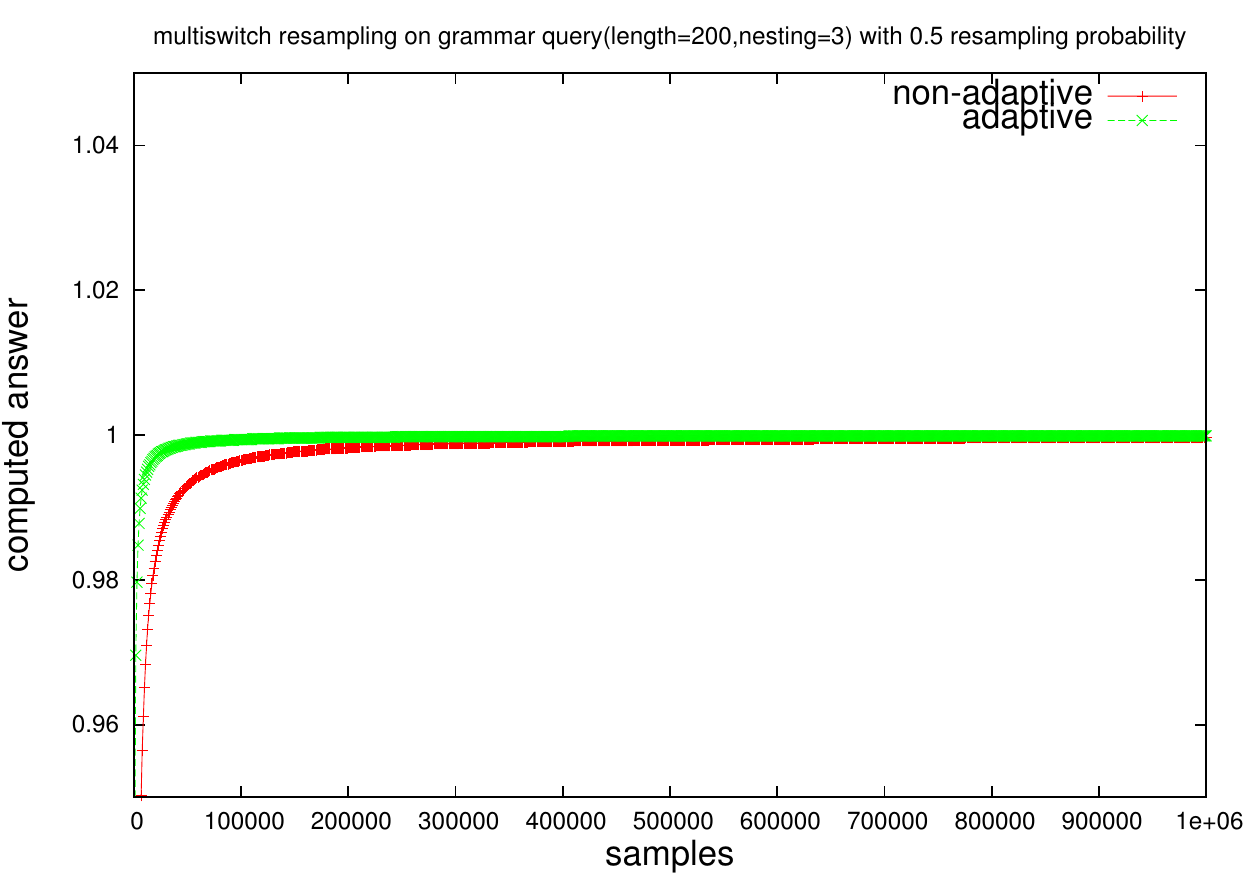}
\end{tabular}
&
  \begin{tabular}[t]{l}
\includegraphics[width=2.75in]{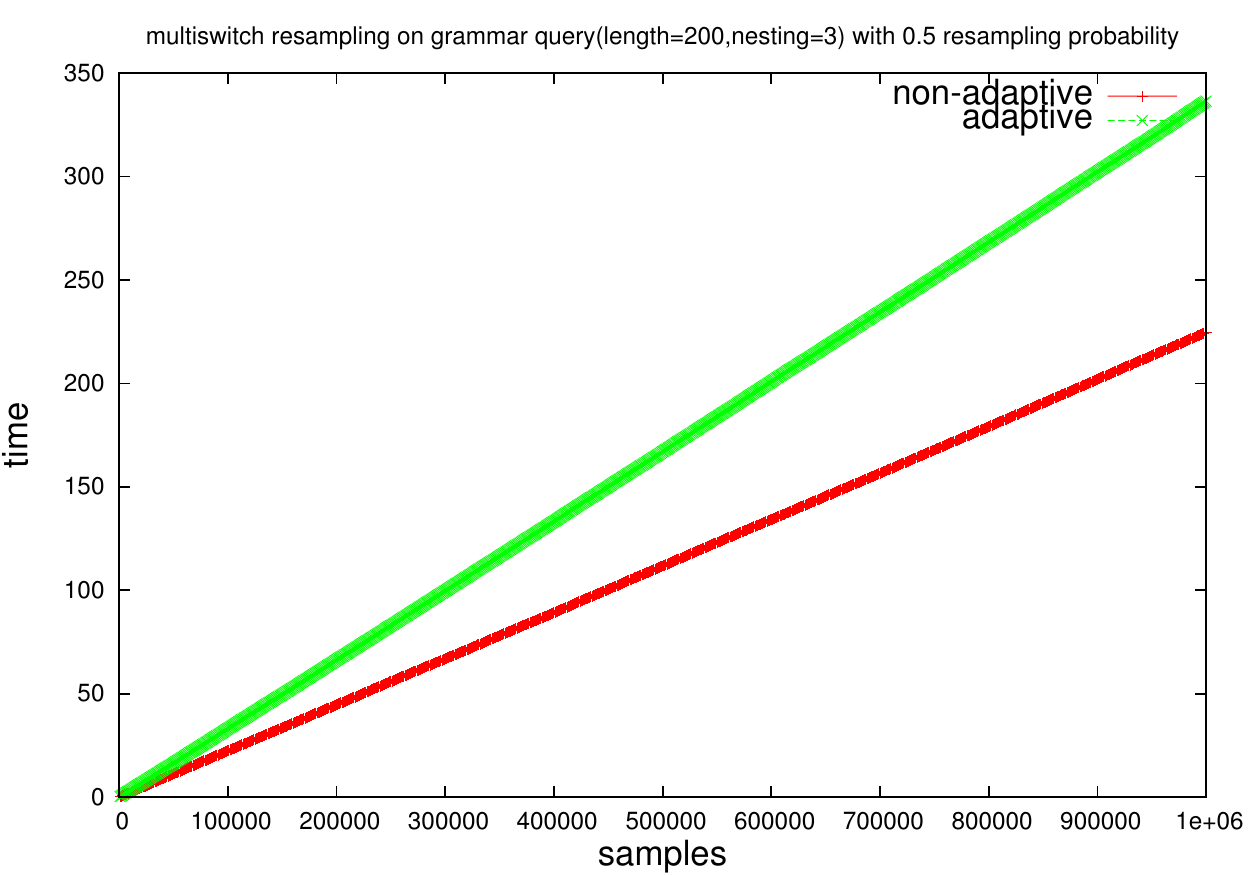}
\end{tabular}\\
(a) & (b)\\
\end{tabular}
\end{center}
\caption{Nesting level in strings with balanced parenthesis.  (a) Condition probability computed
and (b) Running times as functions of sample size.}
\label{fig:grammaranswer}
\end{figure}

\comment{
\begin{figure}
\begin{center}
\includegraphics{Data/hamming/grammar200-3-multi50.eps}
\end{center}
\caption{comparison of answers for grammar test-case}
\label{fig:grammaranswer}
\end{figure}

\begin{figure}
\begin{center}
\includegraphics{Data/hamming/grammar200-3-multi50-time.eps}
\end{center}
\caption{comparison of times for grammar test-case}
\label{fig:grammartime}
\end{figure}

\begin{figure}
\begin{center}
\includegraphics{Data/hamming/grammar200-3-multi50-rejections.eps}
\end{center}
\caption{comparison of cumulative rejections for grammar test-case}
\label{fig:grammarrejections}
\end{figure}
}

\paragraph{\textbf{Reach.}}
The final set of examples are reachability queries in
probabilistic acyclic graphs, of the form shown in
Fig.~\ref{fig:intro-example}.  For the graph shown in Introduction, while
computing  $\cond{\mathtt{reach(a,d)}}{\mathtt{reach(a,e)}}$, the
non-adaptive sampler rejects $8\%$ of the samples, while the
non-adaptive one rejects $1.5\%$.  Similar rejection rates were observed
for larger randomly generated graphs as well.  However, since the
rejection rate of the non-adaptive sampler is low, there is no
significant difference between the convergence of adaptive and
non-adaptive samplers.

\comment{
\begin{figure}
\begin{center}
\includegraphics{Data/reachability/reach810-gibbs.eps}
\end{center}
\caption{comparison of answers for reachability test-case}
\label{fig:reachanswer}
\end{figure}

\begin{figure}
\begin{center}
\includegraphics{Data/reachability/reach810-gibbs-time.eps}
\end{center}
\caption{comparison of times for reachability test-case}
\label{fig:reachtime}
\end{figure}

\begin{figure}
\begin{center}
\includegraphics{Data/reachability/reach810-gibbs-rejections.eps}
\end{center}
\caption{comparison of cumulative rejections for reachability test-case}
\label{fig:reachreject}
\end{figure}

\begin{figure}
\begin{center}
\includegraphics{Data/bnets/test66-multiX-on.eps}
\end{center}
\caption{effect of resampling probability on convergence for multi switch
  resampling} 
\label{fig:resampleprob}
\end{figure}
}

\section{Discussion}
\label{sec:disc}

\label{sec:related}
\comment{
In general the exact inference algorithms in various probabilistic logic
programming languages (PLPs) operate by computing the probability of a
disjunctive normal form (DNF) formula. The disjuncts of the DNF formula
correspond to the different proofs of the goal. Each disjunct is a conjunction
of the probabilistic facts used in that proof \cite{raedt2013probabilistic}.

Exact inference in PRISM \cite{sato2001parameter} tries to reduce the complexity
of this operation by tabling shared subgoals and their explanations. It uses
OLDT search to find all the proofs of a goal. For each proof of a tabled goal,
an explanation is created (an explanation is a conjunction of \emph{msw} atoms
and other tabled goals). These explanations are arranged in the form of an
\emph{explanation graph}. By assuming that explanations of lower level goals do
not depend on higher level goals (\emph{acyclic support} condition), the
probability of the top level goal is computed bottom up using dynamic
programming. The time and space complexity of the algorithm is linear in the
size of the explanation graph. The exact inference algorithm in ProbLog
\cite{de2007problog} uses binary decision diagram (BDD) to represent the DNF
formula. It does this to make the different explanations mutually
exclusive. PRISM avoids this by assuming mutual exclusion of
explanations. However, in these and other languages the difficulty of computing
the probability of a large DNF remains.
}

Sampling-based approximate inference algorithms were proposed by
\citet{cussens2000stochastic} for stochastic logic programs (SLP)s
\cite{muggleton1996stochastic}.  The algorithm defines an MCMC kernel on the
derivations in an SLP.  The technique used to propose the next state (i.e.,
derivation) involves backtracking one step at a time, stopping
with a fixed probability given as a parameter.  Once the backtracking stops, an alternate branch is
sampled and resolution is continued to give next state.  Our
single-switch resampling technique different in that a single \emph{msw} atom is chosen (uniformly at
random) from the state and resampled.   At a more fundamental level,
our sampling technique is largely independent of the query evaluation
process itself.  

An MCMC technique has been proposed for Problog by
\citet{moldovan2013mcmc}. It samples from explanations and
makes use of a special algorithm to make the samples mutually exclusive.
This incurs memory overhead (keeping track of possible worlds used for
each explanation) as well as time overhead (to look back in a chain
for previous uses of the same explanation.  In contrast, we use
Prolog-style evaluation to assure that the samples are pair-wise
mutually exclusive.  \comment{The authors make use of two
parameters which control the reuse of probabilistic choices made in the current
state and the choice of the predicate clauses.  They claim that if small changes
are made to a consistent sample, then it will likely result in another
consistent sample.  In contrast our approach involves explicit adaption of the
proposal distribution to reduce rejections.
}

Adaptive sequential rejection sampling proposed by
\citet{mansinghka2009exact} is an algorithm that explicitly adapts its
proposal for generating samples for high dimensional graphical models.
This algorithm requires the availability of a suitable factorization
of the distribution with logarithmically few dependencies.  Exact
samples over a small set of variables are extended to exact samples
over larger set of variables.  The adaptation scheme described
requires the complete knowledge of the factors in the distribution.
Since PRISM programs represent logical as well as statistical
knowledge, explicit knowledge may not even be available in our case.
Consequently, our work does not rely on an explicit knowledge of
factors.


We presented a MCMC technique for probabilistic logic programs that is
largely independent of the manner in which queries are evaluated in
the underlying logic programming systems.  We defined an adaptive MCMC
algorithm that adapts the probability distribution of individual
switches and their instances to effectively explore the states of the
Markov chain that are consistent with given evidence.  We identified
conditions under which a similar adaptation can be performed to
enable independent samplers to draw more samples that are consistent
with evidence.  Preliminary experiments have shown both the potential
and the limitations of this technique.   

This paper focused on a generic MCMC method and adaptation, and did
not consider the effect of resampling strategies.  The order in which
random processes are sampled may affect the convergence and hence the
quality of inference.  For instance, Decayed
MCMC~\cite{Marthi:etal:UAI02} samples processes based on a temporal
order (resampling more recent processes more frequently).  As future
work, we plan extend our sampler to use an order based on programmer
annotation; whether such annotations can be inferred from the program
is an open problem.  Finally, while sampling-based inference may be
generally deployed, exact inference may still be feasible for queries
with short derivations.  Hence, an interesting direction of future
work is to develop a hybrid inference technique that can combine exact
and approximate inference based on programmer annotation.  Such an
inference technique can be seen as an analogue of the
Rao-Blackwellized Particle Filtering method developed for Dynamic
Bayesian Networks~\cite{Doucet:etal:UAI00}.


\bibliographystyle{acmtrans}

\newpage

\appendix
\section{Acceptance probability computation for MH sampler}
Assume that the Markov chain is in state corresponding to assignment $\sigma$
and a proposal is made to transition to a different state corresponding to
assignment $\sigma'$. The set of switch/instance pairs in $\sigma$ can be
divided into three disjoint sets: $\sigma_{1} = \{(s,i) \vert \sigma'(s,i) =
\bot \}$, $\sigma_{2} = \{(s,i) \vert \sigma'(s,i) \neq \bot \text{ and }
\sigma(s,i) \neq \sigma'(s,i) \}$ and $\sigma_{3} = \{(s,i) \vert \sigma(s,i) =
\sigma'(s,i)\}$. The set of switch/instance pairs in $\sigma'$ can be divided
similarly. The probability of an assignment is simply the product of the
probabilities of the outcomes of all switch/instance pairs in that
assignment. The probability of an assignment $\sigma$ is denoted by
$P(\sigma)$. We denote the number of switch/instance pairs in an
assignment/sub-assignment $\sigma$ by $\vert \sigma \vert$. The acceptance
probability for single switch non-adaptive resampling is 
\[
\frac{P(\sigma')P(\sigma_{1})P(\sigma_{2})1/\vert \sigma'
  \vert}{P(\sigma)P(\sigma'_{1})P(\sigma'_{2})1/\vert \sigma \vert} =
\frac{P(\sigma'_{1})P(\sigma'_{2})P(\sigma'_{3})P(\sigma_{1})P(\sigma_{2})1/\vert
  \sigma'\vert}{P(\sigma_{1})P(\sigma_{2})P(\sigma_{3})P(\sigma'_{1})
  P(\sigma'_{2})1/\vert \sigma \vert} = \frac{\vert \sigma \vert}{\vert \sigma'
  \vert}
\]
Let the probability for forgetting switch in multi switch resampling be $p$. The
acceptance probability can be computed as follows

\[
\frac{P(\sigma')p^{\vert \sigma_{2} \vert}P(\sigma_{2})P(\sigma_{1})\prod_{(s,i)
    \in \sigma_{3}}(1-p+pP(\sigma_{3}(s,i)))}{P(\sigma)p^{\vert \sigma'_{2}
    \vert}P(\sigma'_{2})P(\sigma'_{1})\prod_{(s',i') \in
    \sigma'_{3}}(1-p+p(\sigma'_{3}(s',i')))} =
1
\]
In the case of adaptive sampling such simplifications are not possible. Let the
adapted distribution be denoted by $P'$. The probability of transition from
$\sigma$ to $\sigma'$ in case of single switch resampling is
$P'(\sigma'_{1})P'(\sigma'_{2})/\vert \sigma \vert$. The acceptance
probability is therefore computed as 
\[
\frac{P(\sigma')P'(\sigma_{1})P'(\sigma_{2})1/\vert \sigma'
  \vert}{P(\sigma)P'(\sigma'_{1})P'(\sigma'_{2})1/\vert \sigma \vert}=
\frac{P(\sigma'_{1})P(\sigma'_{2})P'(\sigma_{1})P'(\sigma_{2})1/\vert \sigma'
  \vert}{P(\sigma_{1})P(\sigma_{2})P'(\sigma'_{1})P'(\sigma'_{2})1/\vert \sigma \vert}
\]
Finally we can show that the acceptance probability for multi switch resampling
is
\[
\frac{P(\sigma')P'(\sigma_{1})P'(\sigma_{2})}{P(\sigma)P'(\sigma'_{1})P'(\sigma'_{2})}=
\frac{P(\sigma'_{1})P(\sigma'_{2})P'(\sigma_{1})P'(\sigma_{2})}{P(\sigma_{1})P(\sigma_{2})P'(\sigma'_{1})P'(\sigma'_{2})}
\]

\end{document}